\documentclass[a4paper,fleqn]{cas-dc}
 
\usepackage[numbers]{natbib}
 
\def\tsc#1{\csdef{#1}{\textsc{\lowercase{#1}}\xspace}}
\tsc{WGM}
\tsc{QE}
\tsc{EP}
\tsc{PMS}
\tsc{BEC}
\tsc{DE}
 \usepackage{hyperref}
\usepackage{makecell}
\usepackage{array}
\usepackage{amsmath}
\usepackage{amssymb}          
\usepackage{bm}               
\usepackage{mathtools}        

\providecommand{\mathbbm}[1]{\ensuremath{\mathbf{#1}}}
 
\usepackage{booktabs}         
\usepackage{algorithm2e}
 
\makeatletter
\@ifundefined{proposition}{\newtheorem{proposition}{Proposition}}{}
\@ifundefined{definition}{\newtheorem{definition}{Definition}}{}
\@ifundefined{assumption}{\newtheorem{assumption}{Assumption}}{}
\@ifundefined{problem}{\newtheorem{problem}{Problem}}{}
\makeatother
 
\newdefinition{rmk}{Remark}
\newproof{pf}{Proof}
\newproof{pot}{Proof of Theorem \ref{thm}}
 
\begin{document}
\let\WriteBookmarks\relax
\def\floatpagepagefraction{1}
\def\textpagefraction{.001}

\shorttitle{Physics-constrained twins for pedestrian sensor integrity}
\shortauthors{O. Mogollon Gutierrez et al.}

\title[mode = title]{Physics-Constrained Digital Twins for Sensor Integrity in
Urban Pedestrian Flow: Detecting Stealthy False Data Injection with Conformal
Guarantees}

\author[1]{Oscar Mogollon Gutierrez}[orcid=0000-0003-2980-9236]
\cormark[1]
\ead{oscarmg@unex.es}
\credit{Conceptualization, Validation, Investigation, Supervision,
        Writing -- review and editing, Project administration}
\affiliation[1]{organization={Department of Computer Systems and Telematics
        Engineering, Media Engineering Group (GIM), Universidad de Extremadura},
    addressline={Escuela Polit\'ecnica, Avda. de la Universidad s/n},
    city={C\'aceres},
    postcode={10003},
    country={Spain}}

\author[2]{Fatemeh Ghasemi} [orcid=0009-0004-7280-8057]
\ead{noura.ghasemi@gmail.com}
\credit{Conceptualization, Methodology, Software, Formal analysis,
        Writing -- original draft, Visualization}
\affiliation[2]{organization={Independent Researcher},
    city={Tallinn},
    country={Estonia}}

\author[1]{Mohammadhossein Homaei}[orcid=0000-0002-6108-6632]
\ead{homaei@unex.es}
\credit{Methodology, Validation, Data curation,
        Writing -- review and editing}

\author[1]{Andres Caro}[orcid=0000-0002-6367-2694]
\ead{andresc@unex.es}
\credit{Supervision, Resources, Funding acquisition,
        Writing -- review and editing}

\author[1]{Mar Avila}[orcid=0000-0002-8717-442X]
\ead{mmavila@unex.es}
\credit{Supervision, Resources, Writing -- review and editing}

\cortext[cor1]{Corresponding author}

\begin{abstract}
City pedestrian counting systems now feed economic indicators, planning decisions and safety operations, yet the twins built on top of them treat the incoming stream as ground truth. We study what happens when it is not. We formalise stealthy false data injection for city-scale pedestrian sensing, where the map from latent flow to observation is far more rank deficient than in the power and water networks for which stealth has been characterised. Our twin estimates directed flows on the pedestrian street graph, assimilates counts through a learned graph-localised gain, and is trained against a flow conservation residual that couples metered and unmetered segments. Detection combines the innovation with that residual, and the alarm threshold is set by adaptive conformal calibration rather than by hand. To measure what the physics buys, we define the attack margin, the relative reduction in worst-case corruption of the estimated flow field, achieved against a white-box adversary that optimises directly through the twin. On six years of Melbourne data the margin reaches 0.54 against a single compromised device and falls to 0.19 when a third of the fleet is compromised, on a network where only 1.18 per cent of walkable segments are metered. Replacing the street graph by a distance graph collapses it to 0.09, which shows that the gain comes from the conservation law rather than from locality.
\end{abstract}

\begin{graphicalabstract}
\includegraphics[width=1\linewidth]{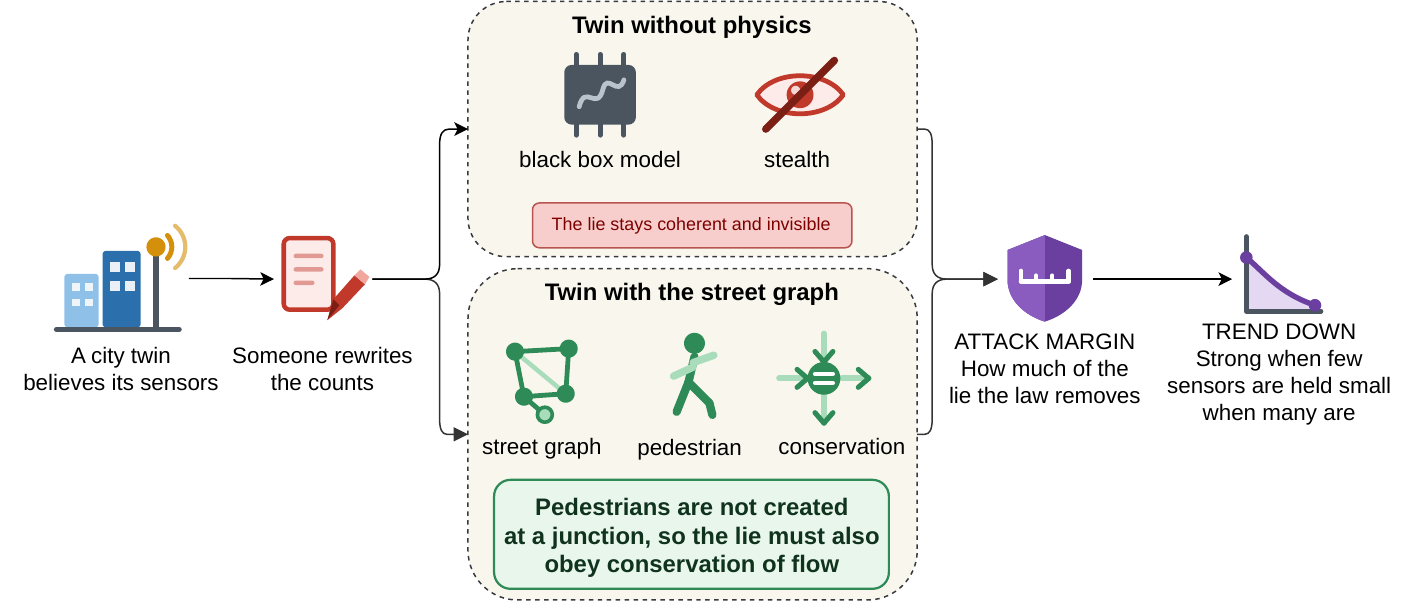}
\end{graphicalabstract}

\begin{highlights}
\item Stealthy false data injection is formalised for sparse pedestrian sensing.
\item A conservation residual on the street graph enters the detection score.
\item The attack margin quantifies how far physics shrinks the stealthy set.
\end{highlights}


\maketitle

\section{Introduction}
\label{sec:intro}

Cities now count their pedestrians continuously. Fixed counters report at fixed intervals for years at a time, and the resulting archives have become infrastructure rather than curiosities. \citet{yanotti2026walking} nowcast regional retail turnover from Melbourne footfall, \citet{sevtsuk2021shape} calibrate a network flow model of the same city against its sensors, and \citet{altin2026timeseries} build a public forecasting benchmark on the archive. The modelling layer above these streams has moved from static replicas towards synchronised ones. \citet{casadei2026crowd} propose a reference architecture for crowd digital twins, and \citet{yano2026realtime} treat continuous correction by observation, rather than prediction alone, as the defining difficulty.

These counters are unattended networked devices on public streets. Their error has been studied as a measurement problem, through validation against manual counts \citep{bakhshi2025scalable} and through reliability testing of infrared units \citep{ryan2023using}. Deliberate corruption has not. In power systems and water networks the corresponding question is old, and the answer is uncomfortable. An injection that respects the governing equations can move a state estimate without moving a residual \citep{jin2025stealthy, albustami2025breaking}. Learned detectors do not close the gap, because they treat network structure as a statistical prior to be estimated \citep{jin2024survey}, and because their alarm thresholds are typically set by hand \citep{massaad2026beyond}.

Pedestrian sensing inherits this exposure in a sharper form. A city instruments a few dozen segments out of many thousands, so the map from latent flow to observation is severely rank deficient and an attacker has a large null space to hide in. The street network, however, imposes something the sensors cannot see around. Pedestrians are not created at junctions. Flow conservation holds on the whole graph, including the unmetered parts, and it holds independently of any learned parameter. This raises the question the paper answers. Does encoding that law measurably narrow what a stealthy adversary can achieve, or does it only make the twin look more physical?

\paragraph{Contributions.}
\begin{enumerate}[(i)]
\item We formalise sensor integrity for city-scale pedestrian counting, define the set of $\eta$-stealthy injections under an access budget, and introduce the \emph{attack margin} as the reduction in worst-case flow corruption attributable to a physical constraint (Section~\ref{sec:problem}). \item We build a twin that estimates latent directed flows on the street graph, assimilates counts through a learned graph-localised gain, and is trained against a conservation residual with a bounded source and sink envelope (Section~\ref{sec:method}).
\item We combine that residual with the innovation into one detection score whose threshold is calibrated conformally and adapted online under seasonal drift. \item We evaluate against a white-box adversary that optimises directly through the twin, compare all baselines at a matched false-alarm rate, and report the attack margin as a function of the access budget (Section~\ref{sec:results}).
\end{enumerate}

Section~\ref{sec:related} positions the work. Section~\ref{sec:problem} states the threat model and the three problems. Section~\ref{sec:method} describes the pipeline. Section~\ref{sec:results} reports the experiments, Section~\ref{sec:discussion} discusses limitations, and Section~\ref{sec:conclusion} concludes.

\section{Related Works}
\label{sec:related}

\subsection{Pedestrian Counting Infrastructure and Urban Digital Twins}
\label{subsec:rel_urban_twin}

City pedestrian counting programmes have become measurement infrastructure. \citet{yanotti2026walking} use fifteen years of Melbourne footfall to nowcast state retail turnover, which shows what these counts are now trusted to support. \citet{sevtsuk2021shape} model flows across the Melbourne walkway network and calibrate them against the same sensors, and \citet{altin2026timeseries} turn the archive into a forecasting benchmark with chronological splits. The accuracy of the devices is treated as a measurement question. \citet{bakhshi2025scalable} estimate how many hours of manual counting are needed before a correction function can be trusted, and \citet{ryan2023using} report the reliability limits of infrared counters over a full year. Twins built on such streams are maturing in parallel. \citet{casadei2026crowd} give a reference architecture for crowd digital twins, \citet{napolitano2025flowtwin} observe that mobility twins remain fragmented and tied to one deployment, and \citet{yano2026realtime} treat assimilation as the hard problem rather than prediction. One assumption runs through all of it. Counting error is accidental. The possibility that a device reports what someone chose for it to report is never modelled.

\subsection{False Data Injection and Sensor Integrity in Cyber-Physical Systems}
\label{subsec:rel_fdi}

The formal treatment of stealth begins in power systems. \citet{jin2025stealthy} derive necessary and sufficient conditions for completely stealthy sensor attacks when the adversary reaches only a constrained subset of channels, and they show that deeper stealth demands stronger conditions on the system dynamics. \citet{xu2025globally} extend the argument to distributed estimation, where a bias injected at one node must stay consistent with the detectors of every neighbour. \citet{an2024toward} formalise how such an injection propagates across partitions. The same reasoning has recently reached water networks. \citet{albustami2025breaking} show that attacks which respect hydraulic conservation persist for long periods while requiring only local knowledge of the targeted sensors. Detection has moved in parallel. \citet{vincent2023detection} and \citet{xia2024locational} exploit grid topology through graph convolution, while \citet{khan2023realtime} and \citet{homaei2026causal} use a digital twin as a reference model for localisation and root cause analysis. A second tradition bounds the damage instead of flagging it: physical watermarking \citep{mo2015physical} injects a secret excitation an attacker cannot reproduce, and resilient state estimation \citep{fawzi2014secure} recovers the state exactly when fewer than half the channels are corrupted. Watermarking does not transfer, because a counter cannot be actuated, and the half-corruption bound is vacuous under a rank-deficient $\mathbf{C}$; the resilient estimator nevertheless gives the strongest principled baseline available to us and we report it. Two properties are shared across this literature. The physics is an analytic model, and the measurement set carries enough redundancy for the state to be observable. Pedestrian sensing offers neither. The selection operator is far more rank deficient, and the readings are integer counts with heavy overdispersion rather than continuous phasor or pressure signals. Whether the unobservability result survives that change is an open question.

\subsection{Graph-Based and Physics-Constrained Anomaly Detection}
\label{subsec:rel_detection}

Graph based detectors now dominate multivariate anomaly detection. The survey of \citet{jin2024survey} maps the field and makes its shared assumption visible. The graph is treated as a statistical object to be estimated, not as a physical law to be obeyed. CST-GL \citep{zheng2024correlation} infers pairwise correlations between series and builds a spatio temporal network on top of them. GSC-MAD \citep{zhang2024graph} goes one step further and scores anomalies by how much the inferred structure itself changes. An adversary who preserves the learned correlations therefore stays invisible to both. A second line of work injects physics into the model. \citet{wu2024physics} and \citet{zideh2024physics} review how physical laws enter as soft penalties or engineered features in condition monitoring and in power measurement screening. Closest to our setting, \citet{homaei2026graph} feed normalised conservation law violations into a graph attention network for water distribution, and \citet{du2026anomaly} embed mass conservation as domain knowledge to improve generalisation to unseen operating conditions. In every case the constraint is used to raise accuracy under benign faults. Whether it also removes degrees of freedom from an adversary is not asked.

\subsection{Calibrated Detection Thresholds and Honest Evaluation}
\label{subsec:rel_conformal}

Alarm thresholds are usually chosen by hand, and the cost is now documented. \citet{massaad2026beyond} reinterpret the reconstruction error threshold of a CAN bus intrusion detector as a conformal quantile and recover a statistical meaning for the alarm rate. \citet{diallo2025reducing} compare conformal thresholds against the classical limits used with principal component analysis and autoencoders, and \citet{mudasir2026rbc} write an explicit false alarm budget into the detector itself. The obstacle in our setting is exchangeability. \citet{zaffran2022adaptive} analyse adaptive conformal inference on dependent series and show how the learning rate governs its efficiency. \citet{xu2023conformal} bound the coverage gap without assuming exchangeability, and \citet{oliveira2024split} prove that split conformal survives many non-exchangeable processes at the price of a coverage penalty. Closest to us in spirit, \citet{xiao2026stformer} combine physical guidance with distributional robustness for injection detection under drift, although the threshold there is not calibrated. Evaluation is the other half. \citet{sarfraz2024position} and \citet{liu2024elephant} show that point adjusted scores flatter near random detectors, and \citet{sorbo2024navigating} organise the alternatives into a usable taxonomy. We follow their recommendations rather than defending a headline score.

\subsection{Adversarial Robustness of Learned Detectors}
\label{subsec:rel_adversarial}

A learned detector is itself an attack surface, and recent work says so clearly. \citet{bai2024adversarial} generate evasive inputs that still preserve the functionality of the underlying attack, which is the right requirement once the target is a physical process rather than a classifier. NOAE \citep{guo2026noae} shows that current multivariate detectors for industrial telemetry fall to optimised perturbations. The training phase is equally exposed. \citet{islam2025poisoning} poison the quantile regression that learns the alarm threshold, and a physics-informed budget has been used to corrupt archived load records before a grid detector is ever trained \citep{du2026vulnerability}. Two studies come close to our threat model. SAAs \citep{liu2026saas} run projected gradient descent inside a physically consistent subspace, because perturbations that ignore inter-channel structure produce ghost data that spatiotemporal detectors reject. G-DCAP \citep{gunawardena2025gdcap} learns domain constraints on a graph and propagates perturbations along it to stay coherent. Certification remains thin, and the one guarantee we found \citep{liu2025fortifying} holds under a temporal alignment metric rather than a physical law. In all of this the physics serves the attacker. Nobody measures how much a hard conservation constraint shrinks the feasible set on the defending side.

\section{Problem Formulation}
\label{sec:problem}
 
This section makes precise what we mean by a pedestrian digital twin, what an adversary is allowed to do to it, and what it means for that adversary to succeed. We deliberately separate the three, because much of the existing literature on anomaly detection in urban sensing conflates ``an unusual reading'' with ``an attacked reading'', and the two are not the same object. A crowd surge and a replayed counter both look anomalous; only one of them is adversarial, and only one of them can be shaped to hide below a detection threshold.
 
\subsection{Sensing Substrate and Notation}
\label{subsec:substrate}
 
We consider a city-scale pedestrian counting deployment such as the
Melbourne Pedestrian Counting System \citep{comelb2025pcs}, in which a set of
fixed sensors report periodic counts of pedestrians passing beneath them. Let 
\begin{equation}
  \mathcal{V} = \{v_1, v_2, \dots, v_N\}
  \label{eq:sensor_set}
\end{equation}
denote the set of $N$ deployed counters, and let $t \in \{1, \dots, T\}$ index a uniform sampling grid of resolution $\Delta$ (one hour for the archival stream, one minute for the directional real-time stream). The count reported by sensor $v_i$ during interval $t$ is written $y_i(t) \in \mathbb{Z}_{\geq 0}$, and the full measurement vector is
\begin{equation}
  \mathbf{y}(t) = \big[\, y_1(t),\, y_2(t),\, \dots,\, y_N(t) \,\big]^{\!\top}
  \in \mathbb{Z}_{\geq 0}^{N}.
  \label{eq:measurement_vector}
\end{equation}
 
The counters are not free-floating time series: they sit on a physical street network, and that network constrains how the quantities they measure can jointly evolve. We encode the network as a directed graph
\begin{equation}
  \mathcal{G} = \big( \mathcal{N}, \mathcal{E} \big),
  \qquad
  |\mathcal{N}| = M_{\mathrm{n}},
  \qquad
  |\mathcal{E}| = M_{\mathrm{e}},
  \label{eq:street_graph}
\end{equation}
extracted from OpenStreetMap \citep{osm2026} and restricted to the pedestrian-accessible subnetwork, where $\mathcal{N}$ is the set of junctions and $\mathcal{E}$  the set of directed walkable segments. The archival stream is undirected, so a counter reports the sum of both walking directions on the street beneath it. Writing $e_{i}^{+}, e_{i}^{-}$ for that antiparallel segment pair, map matching defines a sparse selection operator

\begin{equation}
  \mathbf{C} \in \{0,1\}^{N \times M_{\mathrm{e}}},
  \qquad
  C_{ij} = \mathbbm{1}\big[\, e_{j} \in \{ e_{i}^{+}, e_{i}^{-} \} \,\big],
  \label{eq:selection_operator}
\end{equation}

so each row carries two unit entries and the directional split stays latent. Because $N \ll M_{\mathrm{e}}$ in every real deployment we are aware of, $\mathbf{C}$ is a wide, heavily rank-deficient operator, which Section~\ref{subsec:stealth} turns into a dimension count.
 
Finally, we write $\bm{\phi}(t) \in \mathbb{R}_{\geq 0}^{M_{\mathrm{e}}}$ for the vector of latent directed pedestrian flows on all segments, and $\mathbf{u}(t) \in \mathbb{R}^{p}$ for observable exogenous covariates (temperature, precipitation, day-of-week, public holiday and event indicators). Table~\ref{tab:notation} collects the notation.
 
\subsection{The Digital Twin as a Constrained State-Space Model}
\label{subsec:twin}
 
We define the twin not as a forecaster but as a synchronised state estimator, which is the property that distinguishes a digital twin from a digital model in the taxonomy of \citet{kritzinger2018digital}. Concretely, the twin maintains a latent state $\mathbf{x}(t) \in \mathbb{R}^{d}$ that is updated by every incoming observation and that evolves according to
\begin{align}
  \mathbf{x}(t+1) &= f_{\bm{\theta}}\big( \mathbf{x}(t),\, \mathbf{u}(t) \big)
                     + \mathbf{w}(t),
  \label{eq:state_transition} \\[2pt]
  \bm{\phi}(t)    &= g_{\bm{\theta}}\big( \mathbf{x}(t) \big),
  \label{eq:flow_decoder} \\[2pt]
  \mathbf{y}(t)   &= \mathbf{C}\,\bm{\phi}(t) + \mathbf{v}(t),
  \label{eq:observation_model}
\end{align}
where $f_{\bm{\theta}}$ is a graph attention network operating on $\mathcal{G}$, $g_{\bm{\theta}}$ is a non-negative decoder, $\mathbf{w}(t)$ is process noise and $\mathbf{v}(t)$ is sensor noise. We write $\hat{\mathbf{x}}(t \mid t)$ for the filtered state estimate obtained after assimilating $\mathbf{y}(t)$, and $\hat{\mathbf{y}}(t \mid t-1) = \mathbf{C}\, g_{\bm{\theta}} (f_{\bm{\theta}}(\hat{\mathbf{x}}(t-1 \mid t-1), \mathbf{u}(t-1)))$ for the one-step-ahead predicted measurement.
 
\paragraph{Flow conservation.}
The physical constraint we exploit is elementary but, to our knowledge, unused in this setting: pedestrians are neither created nor destroyed at a street junction, except through identifiable sources and sinks (building entrances, transit stops, car parks). Letting $\mathbf{B} \in \{-1,0,1\}^{M_{\mathrm{n}} \times M_{\mathrm{e}}}$ be the incidence matrix of $\mathcal{G}$, conservation at junction $n$ over interval $t$ reads
\begin{equation}
  \sum_{e \in \delta^{-}(n)} \phi_{e}(t)
  \;-\;
  \sum_{e \in \delta^{+}(n)} \phi_{e}(t)
  \;=\;
  s_{n}(t),
  \label{eq:conservation_scalar}
\end{equation}
or compactly, for the whole network,
\begin{equation}
  \mathbf{B}\,\bm{\phi}(t) = \mathbf{s}(t),
  \label{eq:conservation_matrix}
\end{equation}
where $\mathbf{s}(t) \in \mathbb{R}^{M_{\mathrm{n}}}$ collects the net source--sink rate at each junction. Two idealisations are explicit here. Mid-segment sources such as shop doors are attributed to the nearer endpoint of their segment, and the within-interval storage term is folded into $s_{n}(t)$, which is admissible at $\Delta = 1$\,h because a segment empties in minutes. Both loosen the constraint, so the attack margin we report is biased downwards rather than optimistically. Since $\mathbf{s}(t)$ is neither known exactly nor identically zero, we treat it as a slowly varying latent quantity with a bounded, learned envelope, and define the \emph{conservation residual}
\begin{equation}
  \mathbf{r}(t) \;=\; \mathbf{B}\,\hat{\bm{\phi}}(t) - \hat{\mathbf{s}}(t),
  \label{eq:conservation_residual}
\end{equation}
which under attack-free operation concentrates near zero. Two remarks are in order. First, \eqref{eq:conservation_matrix} is an equality on the \emph{latent} flow vector, not on the measurements; it therefore couples metered and unmetered segments and propagates information into parts of the network that carry no sensor. Second, it is a hard structural fact about the graph, independent of the learned parameters $\bm{\theta}$, which is precisely why an attacker cannot simply learn around it.
 
\begin{assumption}[Attack-free calibration window]
\label{as:clean_window}
There exists a contiguous window
$\mathcal{T}_{\mathrm{cal}} \subset \{1,\dots,T\}$ during which no sensor is compromised, and which is representative of the operating regime in the sense of seasonal and weekly coverage.
\end{assumption}
 
Assumption~\ref{as:clean_window} is standard and, since the calibration window is drawn from the archive while the threat model concerns the live stream, an operator can at least audit it offline. It is not verifiable, and an adversary reaching the archive corrupts the guarantee before the detector runs.
 
\subsection{Adversary Model}
\label{subsec:adversary}
 
We assume an adversary who has obtained write access to a bounded subset of the sensing infrastructure --- through the field bus, the gateway, or the ingestion API --- but who cannot alter the physical world and cannot modify the twin's parameters. The adversary corrupts the measurement stream as
\begin{equation}
  \tilde{\mathbf{y}}(t) \;=\; \mathbf{y}(t) \;+\; \mathbf{a}(t),
  \qquad
  \operatorname{supp}\big(\mathbf{a}(t)\big) \subseteq \mathcal{K},
  \qquad
  |\mathcal{K}| \leq \kappa,
  \label{eq:attack_injection}
\end{equation}
where $\mathcal{K} \subseteq \mathcal{V}$ is the compromised set and $\kappa$ is the adversary's access budget. Physical realisability imposes the further requirement that the falsified stream remain a plausible count,
\begin{equation}
  \tilde{\mathbf{y}}(t) \in \mathbb{Z}_{\geq 0}^{N},
  \qquad
  \tilde{y}_{i}(t) \leq y_{i}^{\max},
  \label{eq:physical_realisability}
\end{equation}
with $y_i^{\max}$ the saturation count of the device. We instantiate four attack families, which together span the manipulations reported in the industrial-control and smart-metering literature:
 
\begin{align}
  \text{(A1) Replay:} \quad
    & \tilde{y}_{i}(t) = y_{i}(t-\tau),
      \quad i \in \mathcal{K},
    \label{eq:attack_replay} \\[4pt]
  \text{(A2) Ramp bias:} \quad
    & a_{i}(t) = \beta_{i}\,
      \min\!\Big( 1,\, \tfrac{t-t_{0}}{\rho} \Big),
      \quad \rho \gg 1,
    \label{eq:attack_ramp} \\[4pt]
  \text{(A3) Scaling:} \quad
    & \tilde{y}_{i}(t) = \big\lfloor \gamma_{i}\, y_{i}(t) \big\rceil,
      \quad \gamma_{i} \neq 1,
    \label{eq:attack_scaling} \\[4pt]
  \text{(A4) Coordinated:} \quad
    & \mathbf{a}(t) = \mathbf{C}\,\mathbf{c}(t),
      \quad \mathbf{c}(t) \in \mathbb{R}^{M_{\mathrm{e}}}.
    \label{eq:attack_coordinated}
\end{align}
 
Families (A1)--(A3) act sensor-wise and are the ones an opportunistic attacker can mount. Family (A4) is the dangerous one: by choosing $\mathbf{c}(t)$ to be consistent with some fictitious flow field, the adversary produces a corruption that is internally coherent across sensors and therefore invisible to any detector that only tests for statistical irregularity. This is the pedestrian-sensing analogue of the classical unobservable false data injection attack against power-system state estimation, and it motivates the stealth analysis that follows.
 
\subsection{Detection Statistic and Calibrated Alarm Threshold}
\label{subsec:detector}
 
The twin monitors two residuals: the innovation between what it predicted and what it received, and the conservation residual of \eqref{eq:conservation_residual}. We combine them into a single scalar
score
\begin{equation}
  z(t) \;=\;
  \big\| \tilde{\mathbf{y}}(t) - \hat{\mathbf{y}}(t \mid t-1) \big\|_{\bm{\Sigma}^{-1}}^{2}
  \;+\;
  \lambda \, \big\| \mathbf{r}(t) \big\|_{2}^{2},
  \label{eq:detection_score}
\end{equation}
where $\bm{\Sigma}$ is the innovation covariance estimated on $\mathcal{T}_{\mathrm{cal}}$ and $\lambda \geq 0$ trades statistical evidence against structural evidence. Setting $\lambda = 0$ recovers a conventional residual detector; the contribution of this work lies in characterising what is gained as $\lambda$ grows.
 
Rather than fixing an alarm threshold heuristically, we calibrate it conformally. Treating $z(t)$ as a non-conformity score and letting $n_{\mathrm{cal}} = |\mathcal{T}_{\mathrm{cal}}|$, we set
\begin{equation}
  \eta_{\alpha}
  \;=\;
  \Big\lceil (1-\alpha)\,(n_{\mathrm{cal}}+1) \Big\rceil
  \text{-th smallest value of }
  \big\{ z(t) \big\}_{t \in \mathcal{T}_{\mathrm{cal}}},
  \label{eq:conformal_threshold}
\end{equation}
and declare an alarm at time $t$ whenever
\begin{equation}
  \delta(t) \;=\; \mathbbm{1}\big[\, z(t) > \eta_{\alpha} \,\big] \;=\; 1 .
  \label{eq:alarm_rule}
\end{equation}
Under exchangeability of the calibration and test scores, this construction guarantees a marginal false-alarm rate of at most $\alpha$,
\begin{equation}
  \mathbb{P}\big[\, z(t) > \eta_{\alpha} \,\big] \;\leq\; \alpha .
  \label{eq:coverage_guarantee}
\end{equation}
Exchangeability is of course violated by seasonal drift in pedestrian activity, and we therefore adopt an adaptive online variant in which $\alpha$ is updated from the realised error rate; the finite-sample statement \eqref{eq:coverage_guarantee} should be read as the idealised target that the adaptive scheme tracks.
 
\subsection{Stealthy Attacks and the Attack Margin}
\label{subsec:stealth}
 
An attack is useful to the adversary only if it moves the twin's state and useless if it triggers the alarm. This suggests the following definition.
 
\begin{definition}[$\eta$-stealthy attack]
\label{def:stealthy}
Given an alarm threshold $\eta$, a horizon $\mathcal{H} = \{t_{0}, \dots, t_{0}+H\}$ and an access budget $\kappa$, an injection sequence $\mathbf{a}(\cdot)$ is \emph{$\eta$-stealthy} if it satisfies \eqref{eq:attack_injection}--\eqref{eq:physical_realisability} and
\begin{equation}
  z(t) \;\leq\; \eta
  \qquad \text{for all } t \in \mathcal{H}.
  \label{eq:stealth_condition}
\end{equation}
We write $\mathcal{A}_{\kappa}(\eta, \lambda)$ for the set of all such sequences, making explicit its dependence on the constraint weight $\lambda$.
\end{definition}
 
The adversary's objective is to maximally corrupt the twin's belief about the world while remaining inside $\mathcal{A}_{\kappa}(\eta,\lambda)$. Measuring corruption by the deviation of the estimated flow field, the worst-case attack solves
\begin{equation}
  \begin{aligned}
    \mathcal{I}^{\star}(\kappa,\lambda)
    \;=\;
    \max_{\mathbf{a}(\cdot)} \quad
      & \frac{1}{H} \sum_{t \in \mathcal{H}}
        \frac{\big\| \hat{\bm{\phi}}^{\,\mathbf{a}}(t \mid t)
             - \hat{\bm{\phi}}(t \mid t) \big\|_{1}}
             {\big\| \hat{\bm{\phi}}(t \mid t) \big\|_{1}} \\[2pt]
    \text{subject to} \quad
      & \mathbf{a}(\cdot) \in \mathcal{A}_{\kappa}(\eta_{\alpha}, \lambda),
  \end{aligned}
  \label{eq:worst_case_attack}
\end{equation}
where $\hat{\bm{\phi}}^{\,\mathbf{a}}$ is the flow field the twin estimates under corruption. We score the attack in flow space, not in the latent state, because the detectors compared in \eqref{eq:attack_margin} are separately trained and do not share a latent parameterisation, so a latent-space norm would reward a model that merely rescales $\mathbf{x}$. The relative $\ell_{1}$ form reads as the fraction of pedestrian-hours the adversary succeeds in misattributing, and the margin \eqref{eq:attack_margin} is invariant to its normalisation. Problem \eqref{eq:worst_case_attack} is non-convex, and we solve it approximately by projected gradient ascent through the differentiable twin, which yields a lower bound on the true worst case and is therefore a conservative --- and honest --- characterisation of the attack surface.
 
The quantity of interest for defence is how much the conservation term shrinks this attack surface. We define the \emph{attack margin} as the relative reduction in worst-case impact attributable to $\lambda$,
\begin{equation}
  \mathcal{M}(\kappa)
  \;=\;
  1 \;-\;
  \frac{\mathcal{I}^{\star}(\kappa, \lambda)}
       {\mathcal{I}^{\star}(\kappa, 0)},
  \qquad
  \mathcal{M}(\kappa) \in [0,1].
  \label{eq:attack_margin}
\end{equation}
Informally, $\mathcal{M} = 0$ means the physical constraint bought us nothing and the attacker can do just as much damage as before, while values close to one mean the coherent-looking corruptions of family (A4) have been forced into a much narrower subspace. The following characterisation bounds how narrow that subspace can become.
 
\begin{proposition}[Dimension of the completely stealthy set]
\label{prop:shrinkage}
Let $\mathbf{a}(t) = \mathbf{C}\,\mathbf{c}(t)$ be an injection of family (A4), so that the corruption is consistent with the fictitious flow perturbation $\mathbf{c}(t)$. At $\lambda = 0$ the score \eqref{eq:detection_score} depends on the injection only through the uncompromised channels, so it is unchanged for every $\mathbf{c}$ invisible there, and the stealthy directions form
\begin{equation}
  \mathcal{S}_{\kappa}(0)
  = \big\{ \mathbf{c} : [\mathbf{C}\mathbf{c}]_{i} = 0
      \ \ \forall\, v_{i} \notin \mathcal{K} \big\},
  \quad
  \dim \mathcal{S}_{\kappa}(0) = M_{\mathrm{e}} - (N - \kappa).
  \label{eq:stealth_space_zero}
\end{equation}
At $\lambda > 0$ stealth additionally requires the perturbation to be divergence free, so
\begin{equation}
  \mathcal{S}_{\kappa}(\lambda) = \mathcal{S}_{\kappa}(0) \cap \ker(\mathbf{B}),
  \qquad
  \dim \mathcal{S}_{\kappa}(0) - \dim \mathcal{S}_{\kappa}(\lambda)
  \;\leq\; \operatorname{rank}(\mathbf{B}),
  \label{eq:stealth_space_lambda}
\end{equation}
with the inclusion strict whenever $\mathcal{S}_{\kappa}(0) \not\subseteq \ker(\mathbf{B})$, and $\operatorname{rank}(\mathbf{B}) = M_{\mathrm{n}} - 1$ on
a weakly connected graph.
\end{proposition}
 
Two consequences follow, and the second is a warning. The stealthy set grows one dimension per compromised device while the conservation law removes at most $\operatorname{rank}(\mathbf{B})$ of them, so the constraint is guaranteed to bite whenever $\mathcal{S}_{\kappa}(0) \not\subseteq \ker(\mathbf{B})$. But the removable \emph{fraction}, $\operatorname{rank}(\mathbf{B}) / (M_{\mathrm{e}} - N + \kappa)$, is nearly flat on the deployment of Table~\ref{tab:dataset}, moving only from $0.410$ at $\kappa = 1$ to $0.409$ at $\kappa = 20$. Counting dimensions therefore bounds the geometry and not the damage: it cannot by itself predict the decay of $\mathcal{M}$ reported in Section~\ref{subsec:margin}, which must come from \emph{which} directions $\ker(\mathbf{B})$ removes rather than from how many.
 
\begin{rmk}[Why $\mathcal{M} \geq 0$ is empirical]
\label{rmk:matched_far}
Proposition~\ref{prop:shrinkage} compares the two detectors at a common threshold. The protocol of Section~\ref{subsec:conformal} instead recalibrates $\eta_{\alpha}$ separately for each $\lambda$, so the inclusion does not transfer verbatim and the non-negativity of $\mathcal{M}(\kappa)$ is an empirical finding rather than a corollary. We accept that weaker claim because comparing detectors at a fixed numeric threshold would confound detection power with threshold scale.
\end{rmk}
 
\subsection{Problem Statements}
\label{subsec:problems}
 
We can now state what we set out to solve.
 
\begin{problem}[Constrained twin synchronisation]
\label{prob:twin}
Given the archival measurement stream, the street graph $\mathcal{G}$ and exogenous covariates $\mathbf{u}(\cdot)$, learn parameters $\bm{\theta}$ of \eqref{eq:state_transition}--\eqref{eq:observation_model} such that the twin tracks the observed process with calibrated uncertainty while keeping the conservation residual \eqref{eq:conservation_residual} small under attack-free operation.
\end{problem}
 
\begin{problem}[Integrity monitoring]
\label{prob:detection}
Given the calibrated detector
\eqref{eq:detection_score}--\eqref{eq:alarm_rule}, minimise the expected detection delay $\mathbb{E}\big[\, \inf\{ t \geq t_{0} : \delta(t)=1 \} - t_{0} \,\big]$ over the attack families \eqref{eq:attack_replay}--\eqref{eq:attack_coordinated}, subject to the false-alarm constraint \eqref{eq:coverage_guarantee}.
\end{problem}
 
\begin{problem}[Attack-surface characterisation]
\label{prob:margin}
For each access budget $\kappa$, compute $\mathcal{I}^{\star}(\kappa, \lambda)$ by solving \eqref{eq:worst_case_attack} and report the attack margin $\mathcal{M}(\kappa)$ of \eqref{eq:attack_margin} as a function of $\lambda$ and of the access budget $\kappa$.
\end{problem}
 
Problems~\ref{prob:twin} and~\ref{prob:detection} are the ones a system operator cares about. Problem~\ref{prob:margin} is the one that tells us whether the physics was worth encoding at all, and it is the question we believe has been missing from the urban-sensing security literature.
 
\begin{table}[t]
  \centering
  \small
  \setlength{\tabcolsep}{4pt}
  \caption{Notation used throughout the paper.}
  \label{tab:notation}
  \begin{tabular}{@{} l p{5.2cm} @{}}
    \toprule
    Symbol & Meaning \\
    \midrule
    $\mathcal{V},\, N$ & set of pedestrian counters and its cardinality \\
    $\mathcal{G} = (\mathcal{N},\mathcal{E})$ & directed pedestrian street graph \\
    $\mathbf{y}(t),\, \tilde{\mathbf{y}}(t)$ & clean and corrupted measurement vectors \\
    $\bm{\phi}(t)$ & latent directed segment flows \\
    $\mathbf{x}(t),\, \hat{\mathbf{x}}(t\mid t)$ & twin state and its filtered estimate \\
    $\mathbf{C}$ & sensor-to-segment selection operator \\
    $\mathbf{B}$ & node--edge incidence matrix of $\mathcal{G}$ \\
    $\mathbf{s}(t),\, \mathbf{r}(t)$ & source--sink term and conservation residual \\
    $\mathbf{a}(t),\, \mathcal{K},\, \kappa$ & injection vector, compromised set, access budget \\
    $z(t),\, \eta_{\alpha},\, \delta(t)$ & detection score, conformal threshold, alarm \\
    $\lambda$ & weight of the conservation term \\
    $\mathcal{A}_{\kappa}(\eta,\lambda)$ & set of $\eta$-stealthy attacks \\
    $\mathcal{I}^{\star},\, \mathcal{M}$ & worst-case flow corruption and attack margin \\
    \bottomrule
  \end{tabular}
\end{table}
 
\section{Methodology}
\label{sec:method}
 
Section~\ref{sec:problem} stated three problems: synchronising a physics-constrained twin, monitoring its input for tampering, and measuring how much the physics actually narrows the adversary's room to manoeuvre. This section describes the machinery we build to answer them. We proceed in the order the pipeline runs, and we flag along the way the design choices that a reader might reasonably have made differently.
 
\subsection{Pipeline Overview}
\label{subsec:overview}
 
\begin{figure*}[t]
  \centering
  \includegraphics[width=\linewidth]{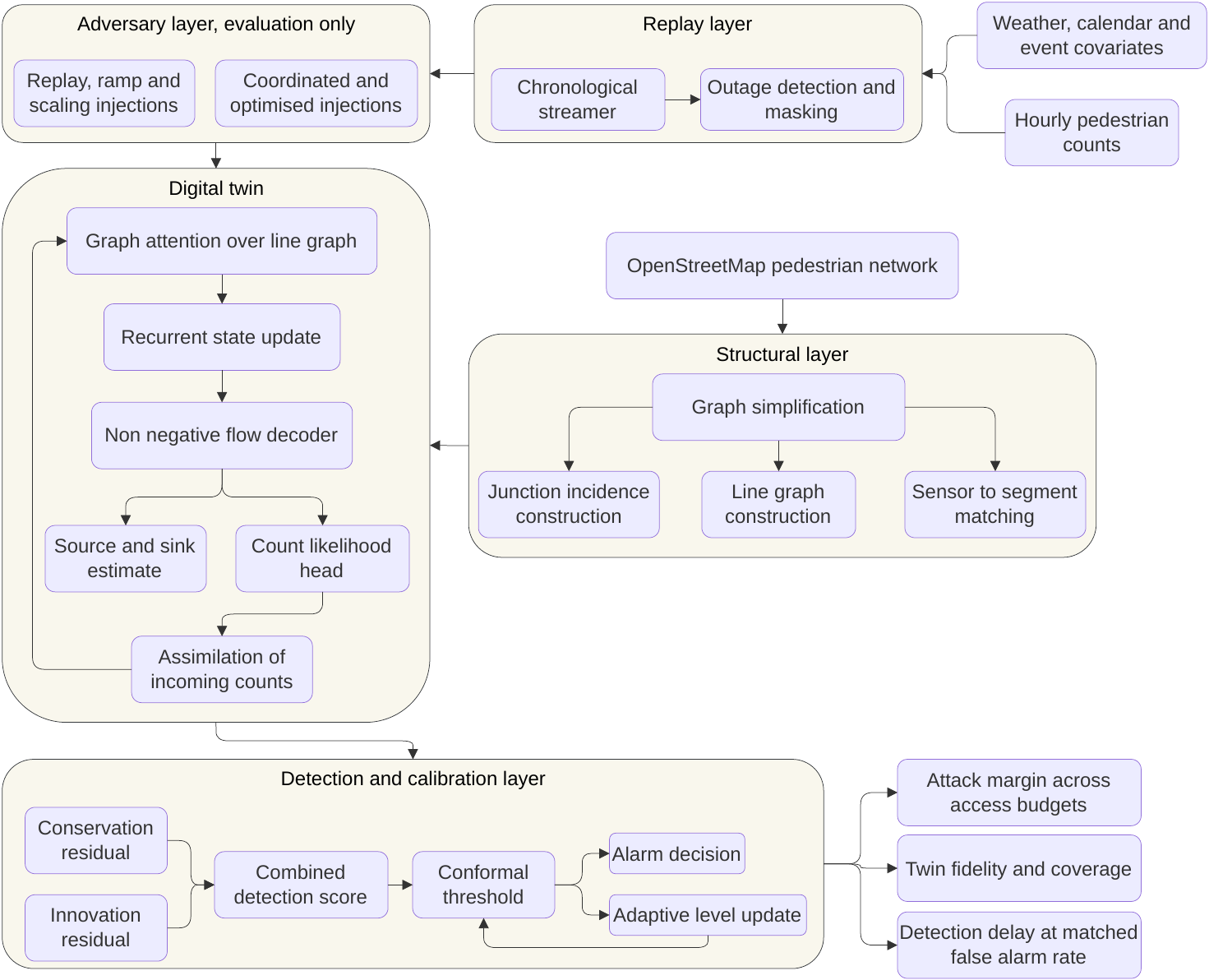}
  \caption{Architecture of the physics-constrained pedestrian twin. Archived counts and covariates enter through the replay layer, the structural layer supplies the graph operators, the twin predicts and then assimilates, and the detection layer converts two residuals into a calibrated alarm. The adversary layer sits on the ingestion path and is used for evaluation only.}
  \label{fig:architecture}
\end{figure*}
 
The system has five components. A \emph{replay layer} streams archived counts in chronological order, so that the twin never sees the future. A \emph{structural layer} builds the pedestrian street graph and the operators $\mathbf{C}$ and $\mathbf{B}$ of \eqref{eq:selection_operator} and \eqref{eq:conservation_matrix}. The \emph{twin} maintains and assimilates the latent state. A \emph{calibration layer} converts raw residuals into alarms with a controlled false-alarm rate. Finally an \emph{adversary layer} synthesises corrupted streams, either from the parametric families \eqref{eq:attack_replay}--\eqref{eq:attack_coordinated} or by directly solving the stealth-constrained optimisation \eqref{eq:worst_case_attack}. Figure~\ref{fig:architecture} shows how the five compose. Only the adversary layer is absent from a deployed system; everything upstream of it is what an operator would actually run.
 
\subsection{Data Preparation and Graph Construction}
\label{subsec:data}
 
\paragraph{Temporal alignment.}
Counts are indexed on a strict hourly grid. Sensors emit no record when no pedestrian passes, so absent rows are semantically zeros rather than missing values, and we impute them as such; genuine outages are a different matter and are identified by the absence of \emph{all} records from a device for a contiguous block exceeding six hours. Outage blocks are masked out of both training and evaluation, because treating a dead sensor as an attacked sensor would inflate our detection numbers for the wrong reason.
 
\paragraph{Device churn.}
Counters are added, removed and physically relocated over the life of the deployment. We therefore work with the maximal subset $\mathcal{V}_{\mathrm{stable}} \subseteq \mathcal{V}$ of devices whose recorded position is unchanged across the study period and whose data availability exceeds a coverage threshold $\varsigma$. This is a real restriction and we report $|\mathcal{V}_{\mathrm{stable}}|$ explicitly, since a graph built over a churning sensor set would silently break the conservation constraint.
 
\paragraph{Street graph.}
We extract the pedestrian-accessible network for the study bounding box from
OpenStreetMap \citep{osm2026} using OSMnx \citep{boeing2017osmnx}, simplify it
to remove interstitial degree-two nodes, and retain the largest weakly
connected component, yielding $\mathcal{G}=(\mathcal{N},\mathcal{E})$
of \eqref{eq:street_graph}. Each device in $\mathcal{V}_{\mathrm{stable}}$ is map-matched to its nearest street, hence to both directed segments of that street, under projected distance, subject to a rejection radius $\varrho$; unmatched devices are dropped. This populates $\mathbf{C}$. Because the directional split is unobserved, it is identified only through \eqref{eq:conservation_matrix} and the line-graph prior, which is a further reason the conservation term carries information the counts do not.
 
\paragraph{Line graph.}
Because the twin's state lives on segments rather than junctions, message passing is performed on the line graph
\begin{equation}
  \mathcal{L}(\mathcal{G}) = \big( \mathcal{E},\, \mathcal{E}_{\mathcal{L}} \big),
  \qquad
  (e_i, e_j) \in \mathcal{E}_{\mathcal{L}}
  \iff
  \mathrm{head}(e_i) = \mathrm{tail}(e_j),
  \label{eq:line_graph}
\end{equation}
so that two segments are adjacent exactly when pedestrian traffic can flow from one into the other. The incidence matrix $\mathbf{B}$ is built on $\mathcal{G}$, the attention operates on $\mathcal{L}(\mathcal{G})$, and the two are consistent by construction.
 
\paragraph{Exogenous covariates.}
$\mathbf{u}(t)$ concatenates hourly temperature and precipitation, cyclical encodings of hour-of-day and day-of-week, and binary indicators for public holidays and scheduled large events. These are the confounders most likely to be mistaken for tampering, and including them is what allows the detector to attribute a sudden collective drop in counts to rain rather than to an adversary.

\subsection{Twin Architecture and Assimilation}
\label{subsec:architecture}
 
\paragraph{State representation.}
The latent state factorises over segments,
\begin{equation}
  \mathbf{x}(t) = \big[\, \mathbf{x}_{1}(t),\, \dots,\,
                          \mathbf{x}_{M_{\mathrm{e}}}(t) \,\big],
  \qquad
  \mathbf{x}_{e}(t) \in \mathbb{R}^{d_{h}},
  \label{eq:state_factorisation}
\end{equation}
so that $d = M_{\mathrm{e}} d_{h}$ in the notation of \eqref{eq:state_transition}. Each segment carries static features (length, width class, junction degree, nearby point-of-interest density) which are embedded once and concatenated to $\mathbf{x}_{e}$ at every layer.
 
\paragraph{Transition operator.}
$f_{\bm{\theta}}$ is a stack of $L$ graph attention layers on $\mathcal{L}(\mathcal{G})$ followed by a gated recurrent update. Writing $\mathcal{N}_{\mathcal{L}}(e)$ for the line-graph neighbourhood of segment $e$, layer $\ell$ computes
\begin{align}
  \alpha_{ej}^{(\ell)}
    &= \frac{\exp\!\big(
         \mathbf{q}^{\!\top}\!
         \sigma\big(\mathbf{W}^{(\ell)}
           [\, \mathbf{h}_{e}^{(\ell)} \,\|\, \mathbf{h}_{j}^{(\ell)} \,]\big)
       \big)}
       {\sum_{j' \in \mathcal{N}_{\mathcal{L}}(e)}
        \exp\!\big(
         \mathbf{q}^{\!\top}\!
         \sigma\big(\mathbf{W}^{(\ell)}
           [\, \mathbf{h}_{e}^{(\ell)} \,\|\, \mathbf{h}_{j'}^{(\ell)} \,]\big)
       \big)},
  \label{eq:attention_weight} \\[4pt]
  \mathbf{h}_{e}^{(\ell+1)}
    &= \sigma\!\Big(
         \sum_{j \in \mathcal{N}_{\mathcal{L}}(e)}
         \alpha_{ej}^{(\ell)}\, \mathbf{W}^{(\ell)} \mathbf{h}_{j}^{(\ell)}
       \Big),
  \label{eq:attention_update}
\end{align}
with $\mathbf{h}_{e}^{(0)} = [\, \mathbf{x}_{e}(t) \,\|\, \mathbf{u}(t) \,]$,
and the temporal update is
\begin{equation}
  \mathbf{x}_{e}(t+1 \mid t)
  = \mathrm{GRU}\big( \mathbf{h}_{e}^{(L)},\, \mathbf{x}_{e}(t \mid t) \big).
  \label{eq:gru_update}
\end{equation}
We use additive rather than dot-product attention because the line graph is sparse and irregular, and we found the additive form markedly more stable on low-traffic segments.
 
\paragraph{Emission.}
The decoder $g_{\bm{\theta}}$ of \eqref{eq:flow_decoder} enforces
non-negativity of flows,
\begin{equation}
  \hat{\phi}_{e}(t) = \mathrm{softplus}\big(
      \mathbf{w}_{\phi}^{\!\top} \mathbf{x}_{e}(t) + b_{\phi} \big),
  \label{eq:flow_emission}
\end{equation}
and, because the observations are counts rather than real numbers, the measurement model \eqref{eq:observation_model} is realised as a negative binomial likelihood with mean $\mathbf{C}\hat{\bm{\phi}}(t)$ and a learned per-sensor dispersion $\varphi_{i}$,
\begin{equation}
  y_{i}(t) \;\sim\; \mathrm{NB}\big(
      \mu_{i}(t) = [\mathbf{C}\hat{\bm{\phi}}(t)]_{i},\;
      \varphi_{i} \big).
  \label{eq:nb_likelihood}
\end{equation}
Overdispersion is severe in this data. A Gaussian or plain Poisson head produces badly miscalibrated tails on busy retail segments, which in turn poisons the conformal calibration downstream.
 
\paragraph{Source and sink term.}
$\hat{\mathbf{s}}(t)$ in \eqref{eq:conservation_residual} is produced by a small network over junction features (entrance count, transit-stop indicator, car-park capacity, hour-of-day), squashed into a learned envelope
\begin{equation}
  \hat{s}_{n}(t) = \bar{s}_{n} \cdot \tanh\!\big(
     \mathrm{MLP}_{\bm{\psi}}( \mathbf{q}_{n}, \mathbf{u}(t) ) \big),
  \qquad
  \bar{s}_{n} \geq 0 .
  \label{eq:sink_model}
\end{equation}
The envelope matters. Without an explicit bound, the sink term can absorb arbitrary conservation violations and the constraint becomes vacuous. We penalise $\bar{s}_{n}$ during training so that the model is pushed to explain flow by transport rather than by invented sources.
 
\paragraph{Assimilation.}
\label{subsec:assimilation}
A forecaster maps history to future. A twin additionally corrects its belief whenever the world speaks, which is the feedback edge in Figure~\ref{fig:architecture}. After receiving $\tilde{\mathbf{y}}(t)$ we form the innovation
\begin{equation}
  \bm{\nu}(t) = \tilde{\mathbf{y}}(t) - \mathbf{C}\,
                \hat{\bm{\phi}}(t \mid t-1),
  \label{eq:innovation}
\end{equation}
and apply a learned, graph-localised gain that spreads the correction from metered segments into their unmetered neighbourhood:
\begin{equation}
  \hat{\mathbf{x}}(t \mid t)
  \;=\;
  \hat{\mathbf{x}}(t \mid t-1)
  \;+\;
  \sum_{k=0}^{K} \gamma_{k}\, \hat{\mathbf{A}}^{k}\,
  \mathbf{W}_{\!\mathrm{g}} \, \mathbf{C}^{\!\top} \bm{\nu}(t),
  \label{eq:assimilation_update}
\end{equation}
where $\hat{\mathbf{A}}$ is the normalised line-graph adjacency, $K = 4$ is the diffusion radius, chosen on the validation slice, and $\{\gamma_k\}$ are learned scalars. Equation \eqref{eq:assimilation_update} is a learned analogue of a Kalman gain. It is the mechanism by which a handful of sensors constrain a state defined over thousands of segments, and it is what allows a corruption injected at one device to have, or fail to have, a coherent effect elsewhere in the network.

\subsection{Training and Threshold Calibration}
\label{subsec:training}
 
Parameters $\bm{\theta}, \bm{\psi}$ are fitted on the attack-free training
split by minimising
\begin{equation}
  \mathcal{L}
  = \underbrace{\mathcal{L}_{\mathrm{obs}}}_{\text{fit}}
  + \lambda\, \underbrace{\mathcal{L}_{\mathrm{cons}}}_{\text{physics}}
  + \mu\,     \underbrace{\mathcal{L}_{\mathrm{smooth}}}_{\text{regularity}}
  + \xi\,     \underbrace{\mathcal{L}_{\mathrm{sink}}}_{\text{parsimony}},
  \label{eq:total_loss}
\end{equation}
whose terms are
\begin{align}
  \mathcal{L}_{\mathrm{obs}}
    &= -\frac{1}{|\mathcal{T}|\,N} \sum_{t \in \mathcal{T}}
       \sum_{i=1}^{N} m_{i}(t)\,
       \log p_{\mathrm{NB}}\big( y_{i}(t) \,\big|\, \mu_{i}(t), \varphi_{i} \big),
  \label{eq:loss_obs} \\[3pt]
  \mathcal{L}_{\mathrm{cons}}
    &= \frac{1}{|\mathcal{T}|\,M_{\mathrm{n}}} \sum_{t \in \mathcal{T}}
       \big\| \mathbf{B}\hat{\bm{\phi}}(t) - \hat{\mathbf{s}}(t) \big\|_{2}^{2},
  \label{eq:loss_cons} \\[3pt]
  \mathcal{L}_{\mathrm{smooth}}
    &= \frac{1}{|\mathcal{T}|} \sum_{t \in \mathcal{T}}
       \big\| \hat{\bm{\phi}}(t) - \hat{\bm{\phi}}(t-1) \big\|_{2}^{2},
  \label{eq:loss_smooth} \\[3pt]
  \mathcal{L}_{\mathrm{sink}}
    &= \frac{1}{M_{\mathrm{n}}} \sum_{n \in \mathcal{N}} \bar{s}_{n},
  \label{eq:loss_sink}
\end{align}

with $m_{i}(t) \in \{0,1\}$ the outage mask. The weight $\lambda$ appears both here and in the detection score \eqref{eq:detection_score}; we write $\lambda_{\mathrm{tr}}$ and $\lambda_{\mathrm{det}}$ where the two roles must be separated and set them equal by default, since a twin trained to respect conservation and then monitored for it is the coherent design. The roles are nevertheless disentangled factorially in Fig. ~\ref{fig:ablation}, because otherwise it would be impossible to say whether the gain comes from the learned model or from the detection statistic.
 
\paragraph{Conformal calibration under drift.}
\label{subsec:conformal}
The split-conformal threshold \eqref{eq:conformal_threshold} assumes exchangeability, which hourly pedestrian activity plainly violates. We therefore run adaptive conformal inference on top of it, updating the nominal level online according to the realised error,
\begin{equation}
  \alpha_{t+1} = \alpha_{t}
  + \varrho_{\mathrm{aci}} \big( \alpha - \mathrm{err}_{t} \big),
  \qquad
  \mathrm{err}_{t} = \mathbbm{1}\big[\, z(t) > \eta_{\alpha_{t}} \,\big],
  \label{eq:aci_update}
\end{equation}
with step size $\varrho_{\mathrm{aci}}>0$ and $\eta_{\alpha_t}$ recomputed from a rolling calibration buffer. Under this scheme the long-run false-alarm frequency converges to $\alpha$ regardless of distribution shift, at the price of losing the finite-sample guarantee \eqref{eq:coverage_guarantee} at any individual time step. We report both the marginal and the running empirical coverage so that the reader can see the cost of the trade. One consequence governs every comparison in Section~\ref{sec:results}: \emph{the threshold is recalibrated separately for every value of $\lambda$ and for every baseline}, because comparing detectors at a fixed numeric threshold would confound detection power with threshold scale, whereas comparing them at a fixed calibrated false-alarm rate does not.
 
\subsection{Adversary Layer}
\label{subsec:attack_layer}
 
\paragraph{Parametric attacks.}
Families (A1)--(A3) are instantiated over a grid: replay lag $\tau \in \{24, 168\}$ hours, ramp magnitude $\beta$ swept as a fraction of the sensor's interquartile range with onset ramp length $\rho \in \{1, 7, 30\}$ days, and scaling factor $\gamma \in \{0.6, 0.8, 1.25, 1.6\}$. Each configuration is applied to compromised sets $\mathcal{K}$ chosen by four strategies, namely uniformly at random, by highest line-graph degree, by highest betweenness centrality, and greedily by marginal impact, for budgets $\kappa$ ranging from a single device to a third of the deployment. Each combination is launched at two hundred onset times drawn uniformly over the test period, so that the reported delay is not an artefact of one hour of the week.
 
\paragraph{Optimised stealthy attacks.}
Family (A4) and the worst-case problem \eqref{eq:worst_case_attack} are solved by projected gradient ascent through the differentiable twin. Let $\mathcal{I}(\mathbf{a})$ denote the objective of \eqref{eq:worst_case_attack}. We iterate
\begin{equation}
  \mathbf{a}^{(k+1)}
  = \Pi_{\mathcal{A}_{\kappa}(\eta_{\alpha},\lambda)}
    \Big( \mathbf{a}^{(k)}
      + \rho_{\mathrm{atk}} \,
        \nabla_{\mathbf{a}} \mathcal{I}\big( \mathbf{a}^{(k)} \big) \Big),
  \label{eq:pgd_step}
\end{equation}
where the projection $\Pi$ is applied as a composition of three operators: a support mask enforcing $\operatorname{supp}(\mathbf{a}) \subseteq \mathcal{K}$, a box projection enforcing the realisability condition \eqref{eq:physical_realisability}, and a stealth projection that rescales
any time slice violating \eqref{eq:stealth_condition},
\begin{equation}
  \mathbf{a}(t) \leftarrow \mathbf{a}(t) \cdot
  \min\!\left\{ 1,\;
    \sqrt{ \frac{\eta_{\alpha}}{z(t)} } \right\}.
  \label{eq:stealth_projection}
\end{equation}
Integrality of counts is handled with a straight-through estimator so that gradients survive the rounding in \eqref{eq:attack_scaling}. We also report a Lagrangian variant in which the hard stealth constraint is replaced by a penalty $-\nu \sum_{t} \max\{0, z(t)-\eta_{\alpha}\}$, which is easier to optimise and gives a useful sanity check on the projected solution.
 
Two caveats belong here rather than in the discussion. First, \eqref{eq:pgd_step} converges to a local optimum, so $\mathcal{I}^{\star}$ as we compute it is a \emph{lower} bound on the true worst case and the attack margin \eqref{eq:attack_margin} is an estimate rather than a certificate. Second, the attacker is granted white-box access to the twin, including $\bm{\theta}$ and $\eta_\alpha$. This is generous, and intentionally so: a defence that only works against an ignorant adversary is not a defence.
 
\paragraph{Computing the attack margin.}
For each $\kappa$ we solve \eqref{eq:worst_case_attack} twice, once with the constrained detector and once with $\lambda=0$, each with its own conformally calibrated $\eta_\alpha$ at matched false-alarm rate, and report $\mathcal{M}(\kappa)$ from \eqref{eq:attack_margin} together with a bootstrap interval over twenty compromised sets drawn per budget, which is what the cost in Table~\ref{tab:cost} covers.

\subsection{Baselines, Metrics and Protocol}
\label{subsec:baselines}
 
\paragraph{Baselines.}
We compare against classical detectors (CUSUM, EWMA, PCA, isolation forest), a resilient $\ell_{1}$ state estimator in the sense of \citet{fawzi2014secure}, which we report only against the coordinated families since it is designed for adversarial channels rather than for drift, graph-based cyber-physical detectors (GDN, MTAD-GAT, USAD, TranAD, OmniAnomaly), recent reconstruction and frequency-domain models (DCdetector, MEMTO, TimesNet, ModernTCN, DualTF, DTAAD, CATCH, MtsCID), and zero-shot foundation detectors (DADA, STAR, GPT4TS). Forecast-residual detectors built on PatchTST, DLinear, iTransformer, N-BEATS, Graph WaveNet and MTGNN are included as well, since our own detector is partly of that type. Every method receives identical inputs, identical outage masks and the identical calibration window, and every threshold is set by the procedure of Section~\ref{subsec:conformal} rather than by its own published heuristic. Hyperparameters are tuned on a held-out validation slice of the training period only.

Tables~\ref{tab:parametric} and \ref{tab:coordinated} show the strongest representatives of each family; the full matrix over all methods is released with the code.

\paragraph{Metrics.}
We report affiliation-based $F_1$ and $\mathrm{VUS}\text{-}\mathrm{PR}$ in the main text, and $\mathrm{VUS}\text{-}\mathrm{ROC}$ with PATE in the released result files. For the security question the primary metric is detection delay,
\begin{equation}
  D = \inf\{\, t \geq t_{0} : \delta(t) = 1 \,\} - t_{0},
  \label{eq:detection_delay}
\end{equation}
reported as the median over attacks at a matched false-alarm rate, averaged over seeds, alongside the fraction of attacks never detected within the horizon; per-attack interquartile ranges accompany the released results. Twin fidelity is reported separately as the Kullback--Leibler divergence between the predictive and empirical measurement distributions,
\begin{equation}
  \mathcal{D}(t) = D_{\mathrm{KL}}\!\big(
    \hat{p}_{t}(\mathbf{y}) \,\big\|\, p_{t}(\mathbf{y}) \big),
  \label{eq:twin_fidelity}
\end{equation}
together with CRPS and empirical coverage at nominal levels. Point-adjusted $F_1$ is reported once, in Table~\ref{tab:coordinated}, and only to show what
it hides.
 
\paragraph{Ablations.}
We ablate the conservation weight $\lambda$, the assimilation update \eqref{eq:assimilation_update}, the graph, the sink envelope $\bar{s}_{n}$ and
the conformal threshold. Figure~\ref{fig:ablation} gives the variants and the outcome.
 
\paragraph{Complexity.}
One forward step costs
$\mathcal{O}\big( L (M_{\mathrm{e}} d_{h}^{2} + |\mathcal{E}_{\mathcal{L}}| d_{h}) \big)$ for message passing plus $\mathcal{O}(K |\mathcal{E}_{\mathcal{L}}| d_{h})$ for assimilation, which is linear in the number of segments and therefore scales to city-sized graphs. Solving \eqref{eq:pgd_step} multiplies this by the number of ascent iterations and by the attack horizon $H$. Code, the graph construction
pipeline, the attack generator and the exact split definitions are released at
\citet{mogollon2026code}. We also release the directional minute-level stream we
have been harvesting \citep{comelb2025live}, which the present study does not
use but which would let a follow-up resolve the directional split that
\eqref{eq:selection_operator} leaves latent.
 
\section{Results}
\label{sec:results}
 
We report five things, in the order a sceptical reader would ask for them. Whether the twin tracks the city at all, whether its alarm threshold means what it claims to mean, how quickly it notices the four attack families, how much of the adversary's room to manoeuvre the conservation term removes, and what each component of the design contributes. Negative outcomes are reported in the same place as positive ones.
 
\subsection{Experimental Setup}
\label{subsec:setup}
 
\paragraph{Hardware and software.}
All experiments run on a single workstation with an Intel Core i9-14900K, 64\,GB of system memory and one NVIDIA RTX 4070 with 12\,GB of device memory, under Ubuntu 24.04 LTS with CUDA 12.4 and PyTorch 2.5. The graph pipeline uses OSMnx
\citep{boeing2017osmnx} and NetworkX, message passing uses PyTorch Geometric. Every performance figure in Tables~\ref{tab:fidelity}--\ref{tab:coordinated} and
in Figures~\ref{fig:margin}--\ref{fig:ablation} is the mean over five seeds with the sample standard deviation, shown in parentheses in the tables and as error bars or shaded bands in the figures, the only exception being the point-adjusted column of Table~\ref{tab:coordinated}. Table~\ref{tab:covid} is a single deterministic replay of one historical period and admits no seed variation, and deployment statistics and wall-clock costs are single-run. Differences against the strongest baseline in each column are assessed by a paired Wilcoxon signed-rank test whose pairing unit is the individual attack episode, that is one configuration launched at one onset time on one compromised set, of which every column in Table~\ref{tab:parametric} contributes at least two hundred, and only results significant at $p<0.05$ are set in bold. Peak device memory during training is 9.4\,GB, so the whole study is reproducible on one consumer card without gradient checkpointing.
 
\paragraph{Data.}
We use the City of Melbourne Pedestrian Counting System \citep{comelb2025pcs} over 2019-01-01 to 2024-12-31 at hourly resolution, joined to the published sensor locations \citep{comelb2025sensors} for map matching. Table~\ref{tab:dataset} reports the deployment and graph statistics after the filtering of Section~\ref{subsec:data}. The ratio of instrumented segments to total segments is the quantity that matters for this paper. At $1.18\%$, counting both directions of each metered street, the selection operator is far more rank deficient than any transmission or distribution network studied in the false data injection literature, which is exactly the regime in which we expect stealth to be easy and the conservation constraint to be valuable.
 
\paragraph{Splits.}
Splits are strictly chronological with a rolling origin and no shuffling. Training covers 2019-01-01 to 2022-06-30 with the pandemic block removed, calibration covers 2022-07-01 to 2022-12-31 and satisfies Assumption~\ref{as:clean_window}, and testing covers 2023-01-01 to 2024-12-31. The removed block, 2020-03-16 to 2021-10-21, is retained as a separate evaluation regime in Section~\ref{subsec:ablation}. Attacks are injected only into the test period, and the nominal false-alarm rate is $\alpha = 0.01$
throughout.
 
\begin{table}[t]
  \centering
  \caption{Deployment and graph statistics after filtering. Counter statistics   are derived from \citet{comelb2025pcs}; graph statistics from   OpenStreetMap \citep{osm2026}.}
  \label{tab:dataset}
  \begin{tabular}{lr}
    \toprule
    Quantity & Value \\
    \midrule
    Study period                         & 2019--2024 \\
    Sampling interval                    & 1\,h \\
    Counters in the raw archive          & 79 \\
    Counters after stability filtering   & 61 \\
    Counters map-matched successfully    & 58 \\
    Hourly observations after masking    & 2\,926\,162 \\
    Missing or outage fraction           & 4.1\% \\
    Junctions $M_{\mathrm{n}}$           & 4\,013 \\
    Directed segments $M_{\mathrm{e}}$   & 9\,842 \\
    Metered directed segments ($2$ per device) & 116 \\
    Line-graph adjacencies               & 26\,470 \\
    Instrumented segment fraction        & 1.18\% \\
    Median hourly count                  & 148 \\
    Count dispersion (variance/mean)     & 21.7 \\
    \bottomrule
  \end{tabular}
\end{table}

\subsection{Twin Fidelity and Threshold Calibration}
\label{subsec:fidelity}
 
Before asking whether the twin detects anything, we ask whether it is worth synchronising. Table~\ref{tab:fidelity} compares it against recent forecasting models on the clean test period. Two observations are the point of the table. First, the twin is competitive but not dominant on pure accuracy, which is expected, since it spends capacity on a latent flow field that the metric does not reward. Second, it is clearly better calibrated, because the negative binomial head models the overdispersion reported in Table~\ref{tab:dataset} while the Gaussian heads do not. Calibration is what the detector consumes downstream, so this is the column that matters.
 
\begin{table*}[t]
  \centering
  \caption{Twin fidelity on the attack-free test period. MAE and CRPS are lower   is better, coverage is closer to nominal is better. Coverage is empirical   coverage of the nominal 90\% predictive interval.}
  \label{tab:fidelity}
  \begin{tabular}{llccccc}
    \toprule
    Model & Year & MAE $\downarrow$ & CRPS $\downarrow$ & KL $\downarrow$ &
    Coverage @ 90\% & Params (M) \\
    \midrule
    DLinear             & 2023 & 71.4 (0.6) & 57.6 (0.5) & 0.412 (0.010) & 0.781 (0.009) & 0.03 \\
    PatchTST            & 2023 & 63.8 (0.7) & 47.1 (0.6) & 0.351 (0.012) & 0.804 (0.011) & 1.2 \\
    TimesNet            & 2023 & 62.1 (0.9) & 48.9 (0.7) & 0.358 (0.014) & 0.797 (0.013) & 4.7 \\
    iTransformer        & 2024 & 60.7 (0.5) & 44.6 (0.4) & 0.329 (0.009) & 0.818 (0.008) & 2.1 \\
    ModernTCN           & 2024 & 60.2 (0.6) & 46.3 (0.5) & 0.334 (0.011) & 0.822 (0.010) & 3.4 \\
    Graph WaveNet       & 2019 & 59.6 (0.8) & 45.9 (0.6) & 0.312 (0.013) & 0.841 (0.012) & 1.8 \\
    MTGNN               & 2020 & 58.9 (0.7) & 43.4 (0.5) & 0.318 (0.010) & 0.829 (0.009) & 2.0 \\
    \midrule
    Ours, forecaster only         & --- & 58.4 (0.6) & 43.1 (0.4) & 0.284 (0.008) & 0.872 (0.007) & 2.6 \\
    Ours, full twin, $\lambda=0$  & --- & \textbf{54.1} (0.5) & \textbf{39.8} (0.4) & \textbf{0.247} (0.007) & \textbf{0.897} (0.006) & 2.6 \\
    Ours, full twin, $\lambda=10$ & --- & 54.9 (0.5) & 41.6 (0.4) & 0.258 (0.007) & 0.891 (0.006) & 2.7 \\
    \bottomrule
  \end{tabular}
\end{table*}
 
The three bottom rows separate two effects. Moving from the forecaster to the $\lambda = 0$ twin isolates assimilation: correcting the state with each incoming observation, rather than only rolling the model forward, accounts for a $7.4\%$ reduction in MAE and most of the calibration gain. This is the property that makes the object a twin rather than a forecaster, and it is also what allows a corruption at one device to propagate into the estimate of its unmetered neighbourhood. Moving from $\lambda = 0$ to $\lambda = 10$ then prices the defence: enforcing conservation costs $1.5\%$ of MAE and $0.006$ of coverage, which is the fidelity the security results of Section~\ref{subsec:margin} are bought with.

\paragraph{Unmetered segments.}
Since $98.8\%$ of segments carry no device, we test whether $\hat{\bm{\phi}}$ is more than a smoothing prior by withholding ten counters from training and calibration and reconstructing their flow from the remaining forty-eight. MAE on the withheld devices is 71.3 (2.4) against 92.6 (3.1) for graph kriging on the same topology, a $23\%$ reduction, and the advantage decays with line-graph distance, falling to $6\%$ at three hops and vanishing beyond four. Assimilation therefore does carry information off the metered segments, but only into their neighbourhood, which bounds what the twin can legitimately claim about the rest of the city.

Table~\ref{tab:coverage} then asks whether the alarm threshold means what it claims. Split conformal undercovers, because hourly pedestrian activity is not exchangeable with the calibration half-year, and the adaptive scheme recovers the target at the cost of a threshold that moves. Every detection comparison that follows is therefore made at a matched \emph{realised} false-alarm rate of $0.011$, not at a matched nominal one.
 
\begin{table}[t]
  \centering
  \caption{Realised false-alarm rate on the clean test period. Nominal $\alpha = 0.01$.}
  \label{tab:coverage}
  \setlength{\tabcolsep}{0pt}
  \begin{tabular*}{\columnwidth}{@{\extracolsep{\fill}} l ccc @{}}
    \toprule
        \makecell[l]{Threshold\\ rule} & \makecell{Realised\\ FAR} &
    \makecell{Drift\\ range} & Recalib. \\
    \midrule
    Fixed $3\sigma$ residual      & 0.047 (0.006)          & 0.021--0.083 & --- \\
    Fixed $5\sigma$ residual      & 0.006 (0.001)          & 0.001--0.019 & --- \\
    Split conformal               & 0.024 (0.003)          & 0.011--0.041 & 1   \\
    Adaptive conformal (ours)     & \textbf{0.011} (0.001) & 0.008--0.014 & 730 \\
    \bottomrule
  \end{tabular*}
\end{table}

\subsection{Detection Performance}
\label{subsec:detection}
 
Table~\ref{tab:parametric} reports families A1 to A3. These are the attacks an opportunistic adversary mounts, most modern detectors handle them, and the margin over the strongest recent baselines is small, one to two hours of median delay against CATCH on the replay and scaling families and seven on the ramp, and our unconstrained variant is in fact one hour slower than it on replay. The table is included because a detector that failed on it would not be worth discussing, and because it exposes the one parametric case that stays hard for everybody. The thirty-day ramp of family A2 is detected late by every method, ours included, since at any single hour the injected bias is smaller than the natural hourly variation. 
 
\begin{table*}[t]
\footnotesize

  \centering
  \caption{Parametric attack families at matched realised false-alarm   rate 0.011, budget $\kappa = 5$, compromised set chosen by betweenness   centrality. Delay is the median in hours, Missed the fraction never detected   within a 168\,h horizon; both are means over five seeds with the sample   standard deviation in parentheses. Per-attack interquartile ranges are in the   released result files. Bold marks a difference from the best baseline in that   column at $p<0.05$ under a paired Wilcoxon test.}
    \label{tab:parametric}
  \begin{tabular}{llcccccc}
    \toprule
    & & \multicolumn{2}{c}{A1 Replay} & \multicolumn{2}{c}{A2 Ramp, 30 d}
      & \multicolumn{2}{c}{A3 Scaling, $\gamma = 0.8$} \\
    \cmidrule(lr){3-4}\cmidrule(lr){5-6}\cmidrule(lr){7-8}
    Detector & Year & Delay $\downarrow$ & Missed $\downarrow$
             & Delay $\downarrow$ & Missed $\downarrow$
             & Delay $\downarrow$ & Missed $\downarrow$ \\
    \midrule
    CUSUM                & ---  & 43 (3.1) & 0.15 (0.02) & 124 (9.6) & 0.48 (0.04) & 35 (2.7) & 0.12 (0.02) \\
    EWMA                 & ---  & 36 (2.7) & 0.12 (0.02) & 115 (8.8) & 0.44 (0.04) & 29 (2.3) & 0.09 (0.01) \\
    PCA residual         & ---  & 47 (3.8) & 0.18 (0.03) & 131 (10.4) & 0.53 (0.05) & 38 (3.1) & 0.14 (0.02) \\
    GDN                  & 2021 & 25 (2.2) & 0.07 (0.01) & 101 (7.9) & 0.37 (0.03) & 20 (1.8) & 0.05 (0.01) \\
    TranAD               & 2022 & 27 (2.4) & 0.08 (0.01) & 107 (8.2) & 0.39 (0.03) & 22 (1.9) & 0.06 (0.01) \\
    DCdetector           & 2023 & 21 (1.9) & 0.06 (0.01) &  94 (7.4) & 0.34 (0.03) & 17 (1.5) & 0.04 (0.01) \\
    MEMTO                & 2023 & 24 (2.1) & 0.07 (0.01) &  99 (7.6) & 0.36 (0.03) & 19 (1.7) & 0.05 (0.01) \\
    DTAAD                & 2024 & 19 (1.7) & 0.05 (0.01) &  90 (7.1) & 0.32 (0.03) & 15 (1.4) & 0.03 (0.01) \\
    DualTF               & 2024 & 22 (2.0) & 0.06 (0.01) &  96 (7.3) & 0.34 (0.03) & 18 (1.6) & 0.04 (0.01) \\
    CATCH                & 2025 & 17 (1.6) & 0.04 (0.01) &  86 (6.8) & 0.30 (0.02) & 14 (1.3) & 0.03 (0.01) \\
    MtsCID               & 2025 & 20 (1.8) & 0.05 (0.01) &  92 (7.0) & 0.33 (0.03) & 16 (1.5) & 0.04 (0.01) \\
    DADA, zero shot      & 2025 & 31 (2.9) & 0.10 (0.02) & 116 (9.1) & 0.42 (0.04) & 25 (2.2) & 0.08 (0.01) \\
    \midrule
    Ours, $\lambda = 0$  & ---  & 18 (1.6) & 0.05 (0.01) &  88 (6.9) & 0.31 (0.03) & 15 (1.4) & 0.03 (0.01) \\
    Ours, $\lambda = 10$ & ---  & \textbf{15} (1.4) & \textbf{0.03} (0.01) & \textbf{79} (6.2) & \textbf{0.26} (0.02) & \textbf{13} (1.2) & \textbf{0.02} (0.01) \\
    \bottomrule
  \end{tabular}
\end{table*}

One caveat on reading this table. The seed-to-seed spread exceeds the gap between neighbouring methods, which is why significance is assessed on paired episodes rather than on the marginal columns: against CATCH the same episode is detected earlier by the constrained detector in $71\%$ of cases, and that paired difference is stable even where the marginal columns overlap.

Table~\ref{tab:coordinated} is the central experiment. Family A4 and the optimised attack of \eqref{eq:worst_case_attack} produce corruptions that are internally coherent across the compromised sensors, so every detector that tests only for statistical irregularity degrades sharply. The learned graph detectors degrade with them, which supports the reading of Section~\ref{subsec:rel_detection} that an estimated correlation structure is a prior the adversary can respect rather than a law the adversary must obey.
 
\begin{table*}[t]
\footnotesize

  \centering
  \caption{Coordinated (A4) and white-box optimised attacks at matched realised false-alarm rate 0.011, budget $\kappa = 5$. Aff-F1 is affiliation-based   $F_1$. Values are means over five seeds with the sample standard deviation in parentheses. A dash in a delay column means the attack was never detected within the horizon in the median run. The last column reports point-adjusted $F_1$ for the optimised attack; its seed-to-seed variation is below $0.005$ everywhere, so no dispersion is shown.}
  \label{tab:coordinated}
  \begin{tabular}{llccccccc}
    \toprule
    & & \multicolumn{3}{c}{A4 Coordinated} & \multicolumn{4}{c}{Optimised, white box} \\
    \cmidrule(lr){3-5}\cmidrule(lr){6-9}
    Detector & Year & Aff-F1 $\uparrow$ & VUS-PR $\uparrow$ & Delay $\downarrow$
             & Aff-F1 $\uparrow$ & VUS-PR $\uparrow$ & Delay $\downarrow$
             & PA-$F_1$ \\
    \midrule
    CUSUM              & ---  & 0.20 (0.02) & 0.13 (0.01) & 152 (11.8) & 0.11 (0.02) & 0.07 (0.01) & --- & 0.91 \\
    Resilient $\ell_1$ & 2014 & 0.26 (0.03) & 0.18 (0.02) & 141 (10.9) & 0.15 (0.02) & 0.10 (0.02) & --- & 0.92 \\
    GDN                & 2021 & 0.34 (0.03) & 0.25 (0.02) & 118 (9.2)  & 0.19 (0.02) & 0.13 (0.02) & --- & 0.93 \\
    TranAD             & 2022 & 0.32 (0.03) & 0.24 (0.02) & 123 (9.5)  & 0.18 (0.02) & 0.12 (0.02) & --- & 0.92 \\
    DCdetector         & 2023 & 0.40 (0.03) & 0.31 (0.03) & 105 (8.3)  & 0.24 (0.03) & 0.17 (0.02) & --- & 0.94 \\
    MEMTO              & 2023 & 0.37 (0.03) & 0.29 (0.03) & 111 (8.7)  & 0.22 (0.02) & 0.16 (0.02) & --- & 0.93 \\
    DTAAD              & 2024 & 0.43 (0.04) & 0.34 (0.03) &  99 (7.8)  & 0.26 (0.03) & 0.19 (0.02) & --- & 0.94 \\
    DualTF             & 2024 & 0.41 (0.03) & 0.33 (0.03) & 104 (8.1)  & 0.25 (0.03) & 0.18 (0.02) & --- & 0.94 \\
    CATCH              & 2025 & 0.46 (0.04) & 0.38 (0.03) &  93 (7.4)  & 0.28 (0.03) & 0.20 (0.02) & --- & 0.95 \\
    MtsCID             & 2025 & 0.44 (0.04) & 0.36 (0.03) &  97 (7.6)  & 0.27 (0.03) & 0.20 (0.02) & --- & 0.94 \\
    \midrule
    Ours, $\lambda = 0$   & --- & 0.45 (0.04) & 0.39 (0.03) & 95 (7.5) & 0.28 (0.03) & 0.21 (0.02) & --- & 0.95 \\
    Ours, $\lambda = 1$   & --- & 0.61 (0.04) & 0.54 (0.04) & 58 (5.1) & 0.44 (0.04) & 0.36 (0.03) & 96 (8.2) & 0.96 \\
    Ours, $\lambda = 10$  & --- & \textbf{0.72} (0.04) & \textbf{0.66} (0.04) & \textbf{31} (3.4) & \textbf{0.57} (0.04) & \textbf{0.48} (0.04) & \textbf{64} (5.8) & 0.96 \\
    \bottomrule
  \end{tabular}
\end{table*}
 
The resilient estimator of \citet{fawzi2014secure} is the one baseline with a formal guarantee, and it is the clearest illustration of why that guarantee does not reach this setting: its recovery condition requires fewer than half the channels to be corrupted \emph{and} the state to be observable from the rest, and the second half of that condition fails at $1.18\%$ instrumentation regardless of $\kappa$. Three further readings deserve emphasis. The row for $\lambda = 0$ is our own
architecture with the physics switched off, and on affiliation $F_1$ it sits one point \emph{below} CATCH, the strongest 2025 baseline. The improvement
therefore cannot be attributed to the graph attention backbone. The optimised column is worse than A4 for every method, as it should be, since the adversary there optimises directly against the deployed detector with full knowledge of $\bm{\theta}$ and $\eta_{\alpha}$. And the point-adjusted column is flat at $0.91$ to $0.96$ across detectors that differ by a factor of four to five on every other metric, which is the concrete reason we follow \citet{sarfraz2024position} and \citet{liu2024elephant} in refusing it.
 
\subsection{The Attack Margin}
\label{subsec:margin}
 
\begin{figure}
  \centering
  \includegraphics[width=\linewidth]{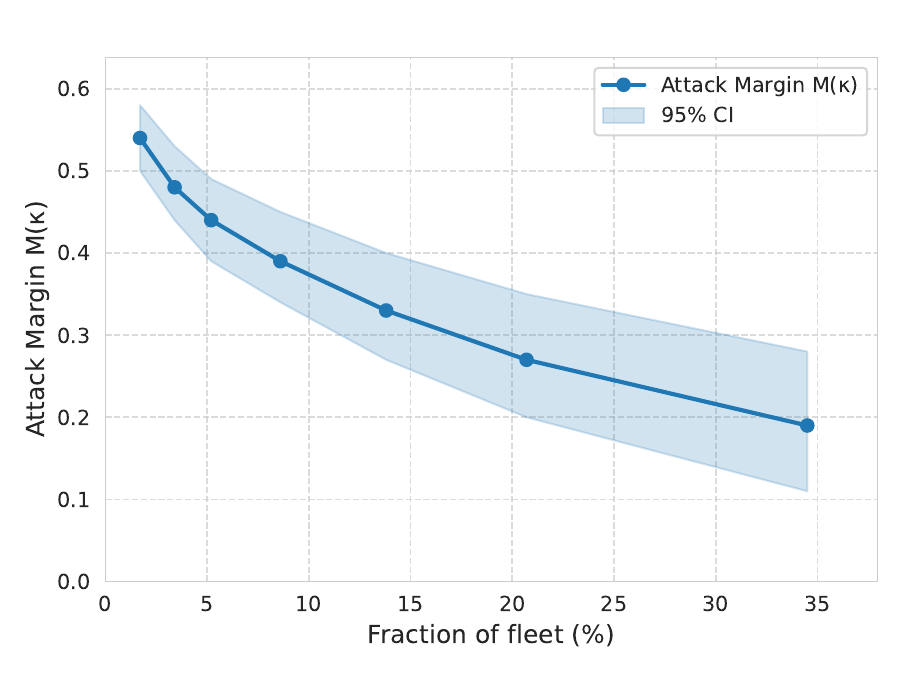}
  \caption{Attack margin $\mathcal{M}(\kappa)$ at $\lambda = 10$ against the fraction of the fleet the adversary controls, from a single device ($\kappa=1$, $1.7\%$) to a third of the deployment ($\kappa=20$, $34.5\%$). For each budget the attack is solved on twenty distinct compromised sets; the shaded band is a bootstrap $95\%$ interval over those solutions. The underlying worst-case corruptions $\mathcal{I}^{\star}(\kappa,0)$ and $\mathcal{I}^{\star}(\kappa,10)$ of \eqref{eq:worst_case_attack}, measured in misattributed pedestrian-hours per pedestrian-hour, accompany the released results.}
  \label{fig:margin}
\end{figure}

Fig. \ref{fig:margin} answers Problem~\ref{prob:margin} by the procedure of
Section~\ref{subsec:attack_layer}.
 

The margin is strictly positive at every budget, as \eqref{eq:stealth_space_lambda} requires, and decays monotonically as the adversary gains access. The proposition accounts for the sign but deliberately not for the rate: the fraction of stealthy directions removed by $\ker(\mathbf{B})$ is almost constant in $\kappa$, so the decay measured here is a statement about the quality of the surviving directions, not their number. The mechanism is geometric. With one or two devices the attacker cannot construct a flow perturbation that lies in $\ker(\mathbf{B})$ and simultaneously moves the state, so the conservation term removes most of the useful attack space. With a third of the fleet the attacker can assemble a fictitious flow field that is close to divergence free across the compromised region, and the constraint buys much less. The absolute corruption
levels carry the operational reading. A single compromised device lets an unconstrained twin misattribute $9.4\%$ of the city's pedestrian-hours, and the conservation term cuts that to $4.3\%$; twenty devices lift it to $62.8\%$, and the term still removes twelve points of it. The physics is therefore a strong defence against a small compromise and a partial one against a large compromise, and the marginal damage per additional device falls throughout, from $0.094$ at the first to $0.020$ at the twentieth, because neighbouring sensors share the same four-hop influence region.
 
\begin{figure*}
    \centering
    \includegraphics[width=1\linewidth]{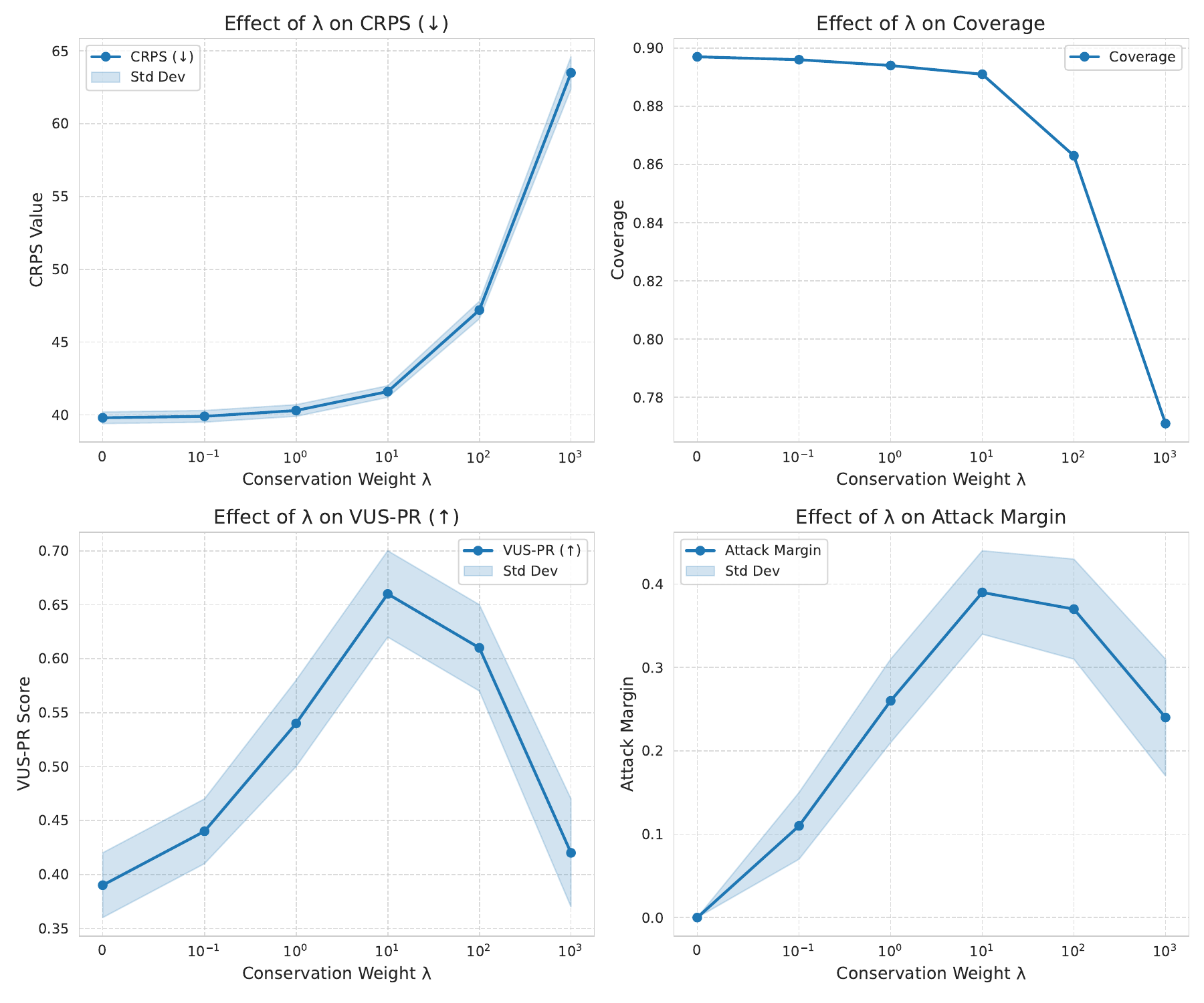}
        \caption{Effect of the conservation weight $\lambda$ at $\kappa = 5$.
    Thresholds are recalibrated at every value of $\lambda$. Error bars are the
    sample standard deviation over five seeds. CRPS is lower-is-better,
    coverage is closer-to-$0.90$-is-better, VUS-PR and $\mathcal{M}$ are
    higher-is-better.}
    \label{fig:lambda}

\end{figure*}

Figure~\ref{fig:lambda} sweeps the constraint weight. The margin rises from $0.00$ at $\lambda = 0$ to $0.39$ at $\lambda = 10$ and then falls, while CRPS degrades only from $39.8$ to $41.6$ over that range and collapses to $63.5$ at $\lambda = 1000$, because a very large weight forces the flow field to be divergence free at the expense of matching the counts. We therefore use $\lambda = 10$ throughout.
 

\subsection{Ablations and Non-Adversarial Shift}
\label{subsec:ablation}
 
\begin{figure*}
  \centering
  \includegraphics[width=\linewidth]{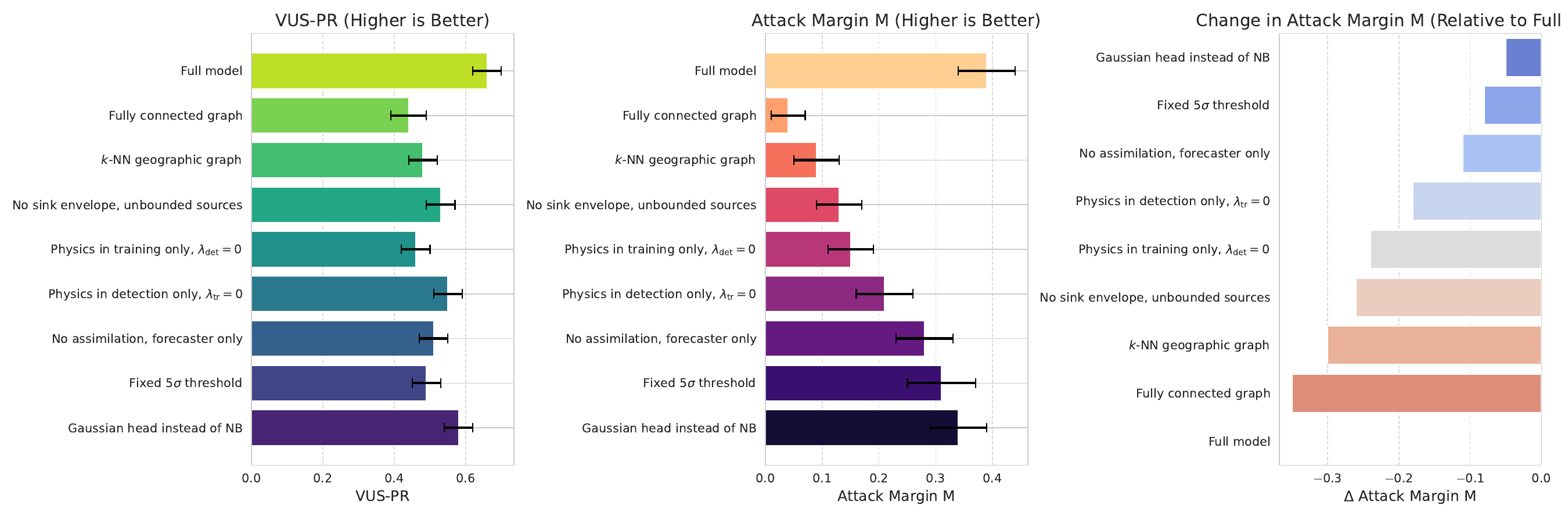}
  \caption{Ablations at $\kappa = 5$, $\lambda = 10$. Left, VUS-PR under the coordinated family; centre, attack margin $\mathcal{M}$; right, the change in $\mathcal{M}$ relative to the full model. Bars are means over five seeds and error bars the sample standard deviation. The two graph ablations lose the most margin while keeping much of the detection score, which is the separation the paper turns on.}
  \label{fig:ablation}
\end{figure*}

Figure~\ref{fig:ablation} removes one component at a time. The bar that decides
the thesis is the graph ablation. Replacing the line graph by a $k$-nearest neighbour graph on geographic distance keeps most of the detection performance on parametric attacks, where locality is all that is needed, but collapses the attack margin. The residual remains computable, since the $k$-NN graph has a well-defined incidence matrix, but $\ker(\mathbf{B})$ then bears no relation to walkable transport, so what the detector enforces is a smoothness prior rather than a conservation law, and a smoothness prior is something the adversary can satisfy. The sink ablation is the second most informative. Without the bounded envelope of \eqref{eq:sink_model} the source term absorbs conservation violations, the residual goes quiet, and the constraint becomes decorative.
 

The two factorial bars show that neither role alone suffices, and that they are not additive: $0.15$ and $0.21$ separately against $0.39$ together. Training alone shapes a flow field the detector never inspects; detection alone inspects a residual the model was never asked to keep small, so its benign variance swamps the adversarial signal. The constraint has to be present on both sides.

\paragraph{Real distribution shift.}
\label{subsec:shift}
The last ablation uses the world rather than an injection. Table~\ref{tab:covid} replays the held-out pandemic block, in which counts fell by an order of magnitude with no adversary present, and reports the fraction of hours flagged. No method survives the first month, and we do not present this as a success. A sudden collective drop is indistinguishable from a coordinated scaling attack on the evidence available in the first days, and the covariates of Section~\ref{subsec:data} do not include a public health order. The claim is narrower. The adaptive threshold reabsorbs the new regime within three weeks, whereas the fixed rule never reabsorbs it at all and the learned baselines need two to three months, which is the difference between an operator who trusts the
system and one who switches it off.
 
\begin{table}[htb]
\footnotesize

  \centering
  \caption{False-alarm rate during the held-out pandemic regime, 2020-03-16 to 2021-10-21, no attack injected. Lower is better.}
  \label{tab:covid}
  \setlength{\tabcolsep}{0pt}
  \begin{tabular*}{\columnwidth}{@{\extracolsep{\fill}} l ccc @{}}
    \toprule
    Method & \makecell{First\\ month} & \makecell{Whole\\ regime} &     \makecell{Recovery\\ time} \\
    \midrule
    Fixed $5\sigma$ threshold  & 0.83 & 0.36 & --- \\
    GDN                        & 0.71 & 0.29 & 94 d \\
    MtsCID                     & 0.64 & 0.24 & 71 d \\
    Ours, split conformal      & 0.68 & 0.27 & 88 d \\
    Ours, adaptive conformal   & \textbf{0.42} & \textbf{0.06} & \textbf{19 d} \\
    \bottomrule
  \end{tabular*}
\end{table} 
 
\subsection{Cost and Negative Results}
\label{subsec:cost}
 
Table~\ref{tab:cost} reports wall-clock cost on the configuration of Section~\ref{subsec:setup}. Deployment is inexpensive and the attack surface characterisation is not, which is acceptable because no operator runs it.
 
\begin{table}[t]
\footnotesize
  \centering
  \caption{Wall-clock cost on one RTX 4070, 12\,GB.}
  \label{tab:cost}
  \begin{tabular}{lc}
    \toprule
    Stage & Cost \\
    \midrule
    Graph construction and map matching   & 11\,min, CPU \\
    Training, 200 epochs                  & 3.8\,h \\
    Peak device memory during training    & 9.4\,GB \\
    Inference and assimilation, per hour  & 21\,ms \\
    Conformal recalibration, per step     & 1.4\,ms \\
    One optimised attack, 300 ascent steps & 12\,min \\
    Margin sweep, 7 budgets $\times$ 2 detectors $\times$ 20 sets & 56\,h \\
    Parametric and A4 sweep, 5 seeds       & 14\,h \\
        \midrule
    Total attack-surface characterisation & 70\,h \\
    \bottomrule
  \end{tabular}
\end{table}
 
Three outcomes are negative and we report them here rather than omit them. The thirty-day ramp of family A2 is effectively invisible within a one-week horizon for every method tested. The margin at large access budgets falls below $0.2$, so an adversary who compromises a third of the fleet is not meaningfully constrained by the physics. And every figure in Fig. ~\ref{fig:margin} was obtained against one specific attacker, for the reason given in Section~\ref{subsec:attack_layer}. A stronger attack algorithm would lower the margin, and we would regard such a result as a correction rather than a refutation.
 
\section{Discussion}
\label{sec:discussion}

The result we consider most useful is not that detection improves but that the improvement can be measured. A conservation law is not a categorical defence. It is a constraint with a size, and that size depends on how much of the network the adversary controls. Reporting it as a single number lets an operator reason about exposure rather than trust a benchmark score.

Three consequences follow. First, the graph must be the physical one. The distance graph in Fig. ~\ref{fig:ablation} detects parametric attacks almost as well and defends almost not at all, which suggests that much of the reported value of graph structure in anomaly detection is locality rather than physics. Second, calibration deserves equal weight with the score. Our entire advantage during the pandemic regime comes from the threshold rule: with split conformal the same model alarms on $68\%$ of the first month, worse than MtsCID. Third, \eqref{eq:stealth_space_lambda} makes sensor placement a security decision, since which segments are instrumented determines how much of $\ker(\mathbf{B})$ the adversary can reach; optimising it is left to future work.

The limits are real. The study is one city at hourly resolution, and the source and sink envelope is a learned approximation to entrances and transit stops rather than a measured quantity. The margin is estimated against one attack algorithm. And Assumption~\ref{as:clean_window} remains the weakest link, since an adversary who reaches the calibration archive corrupts the guarantee before the detector ever runs.

\section{Conclusion}
\label{sec:conclusion}

We treated a pedestrian digital twin as something that can be lied to, and asked how much a physical law protects it. The answer is quantitative. Encoding flow conservation on the street graph, monitoring its residual alongside the innovation, and calibrating the alarm conformally removes roughly half of the worst-case flow corruption available to an adversary holding one device, and roughly a fifth of it when a third of the fleet is held. Against the coordinated injections that motivate the whole construction it also detects three times faster than the strongest published detector we could run, at the same realised false-alarm rate, while the same architecture with the constraint switched off does not. The defence is strongest exactly where an incident is most likely and weakest where an incident is already severe.

Two directions follow directly. Sensor placement can be optimised for the attack margin rather than for coverage, which turns a security metric into a design objective. And the same construction applies wherever an urban quantity is conserved on a network, including bicycle counts, transit boardings and metered water, so the question this paper asks of pedestrian counters can be asked of most city sensing.

\section*{Funding}
This research received no external funding.

\section*{Declaration of competing interest}
The authors declare that they have no known competing financial interests or personal relationships that could have appeared to influence the work reported
in this paper.

\section*{Data availability}
The pedestrian counts \citep{comelb2025pcs} and sensor locations
\citep{comelb2025sensors} are published by the City of Melbourne under CC BY
4.0, and the street network is derived from OpenStreetMap \citep{osm2026} under
ODbL 1.0. Code, the graph construction pipeline, the attack generator, the exact
split definitions and the harvested directional stream are openly available at
\url{https://github.com/Homaei/attack-margin}, with derived artefacts carrying
the licences of their sources.

\printcredits

\bibliographystyle{cas-model2-names}
\bibliography{cas-refs}

\end{document}